%% file: sn-article.tex
\documentclass[pdflatex,sn-mathphys-num]{sn-jnl}
\usepackage{graphicx}
\usepackage{multirow}
\usepackage{amsmath,amssymb,amsfonts}
\usepackage{amsthm}
\usepackage[title]{appendix}
\usepackage{xcolor}
\usepackage{textcomp}
\usepackage{booktabs}
\usepackage{algorithm}
\usepackage{algorithmicx}
\usepackage{algpseudocode}
\usepackage{listings}
\usepackage{tabularx}
\usepackage{hyperref}
\usepackage{rotating}
\hypersetup{hypertexnames=false}


\begin{document}

\title[Benchmarking 3D Medical Vision Foundation Models and Transferability Estimation]{Benchmarking Self-Supervised 3D Medical Vision Foundation Models and Transferability Estimation}

\author[1,2,3]{\fnm{Jiaqi} \sur{Tang}}\email{jiaqi\_tang@pku.edu.cn}
\equalcont{These authors contributed equally to this work.}

\author[1,2,3]{\fnm{Yu} \sur{Wang}}\email{mywork@bupt.edu.cn}
\equalcont{These authors contributed equally to this work.}

\author[1]{\fnm{Yang} \sur{Liu}}\email{yangliu@pku.edu.cn}
\equalcont{These authors contributed equally to this work.}

\author[6]{\fnm{Jing} \sur{Tao}}\email{taojing\_xj@163.com}

\author[4]{\fnm{Fandong} \sur{Zhang}}\email{zhangfandong@deepwise.com}

\author[4]{\fnm{Zhou} \sur{Zhen}}\email{zhouzhen@deepwise.com}

\author[5]{\fnm{Xiahai} \sur{Zhuang}}\email{zxh@fudan.edu.cn}

% The asterisk (*) denotes the corresponding author
\author*[1,2,3]{\fnm{Qingchao} \sur{Chen}}\email{qingchao.chen@pku.edu.cn}

\author*[6,7]{\fnm{Yining} \sur{Yang}}\email{yangyn5126@163.com}

\author[8]{\fnm{Yizhou} \sur{Yu}}\email{yizhouy@acm.org}

% --- Affiliations ---

\affil[1]{\orgdiv{National Institution of Health Data Science}, \orgname{Peking University}, \orgaddress{\city{Beijing}, \country{China}}}

\affil[2]{\orgdiv{Institute of Medical Technology}, \orgname{Peking University}, \orgaddress{\city{Beijing}, \country{China}}}

\affil[3]{\orgdiv{State Key Laboratory of General Artificial Intelligence}, \orgname{Peking University}, \orgaddress{\city{Beijing}, \country{China}}}

\affil[4]{\orgname{Deepwise Co., Ltd.}, \orgaddress{\country{China}}}

\affil[5]{\orgdiv{School of Data Science}, \orgname{Fudan University}, \orgaddress{\city{Shanghai}, \country{China}}}

\affil[6]{\orgdiv{Department of Cardiology}, \orgname{People's Hospital of Xinjiang Uyghur Autonomous Region}, \orgaddress{\city{Urumqi}, \country{China}}}

\affil[7]{\orgname{Xinjiang Key Laboratory of Cardiovascular Homeostasis and Regeneration Research}, \orgaddress{\city{Urumqi}, \country{China}}}

\affil[8]{\orgname{The University of Hong Kong}, \orgaddress{\city{Hong Kong}, \country{China}}}

\abstract{Self-supervised pretraining has expanded the pool of 3D medical foundation models, but published evaluations use different datasets, endpoints and fine-tuning protocols. Models also differ in architecture, pretraining objective and data, while downstream targets range from voxel-level masks to patient-level diagnoses and outcomes. As a result, published findings provide little directly comparable evidence for choosing a model for a new task. Direct comparison of adaptation performance requires fine-tuning candidates on labelled target data, at substantial computational cost. We benchmark 23 candidates across the same 14 segmentation, classification and regression tasks spanning multiple anatomies and modalities. No candidate ranked among the top three across all segmentation tasks or across all classification and regression tasks; a convolutional encoder could outperform a transformer trained with the same objective on one task but underperform it on another. We benchmark transferability estimators against the fine-tuned rankings and find that their agreement varies across dense segmentation, classification and regression. InsightRank uses task-appropriate frozen-feature statistics and architecture-aware pairwise comparisons to rank candidates before fine-tuning. Across the 23-model pool, mean Kendall $\tau$-b was 0.431 for segmentation and 0.352 for classification and regression. The segmentation Top~3 contained a task winner on four of nine tasks; the image-level Top~3 contained none on the five tasks, despite better full-ranking agreement than the matched conventional estimators. A three-model shortlist reduced recorded extraction-plus-training job time by 85.8\% across seven segmentation tasks, excluding untimed statistic computation. Together, the benchmark provides a common reference for model selection and InsightRank offers a lower-cost way to prioritize fine-tuning, with shortlist performance that differs by target type.}


\maketitle

\input{sections/main_text.tex}

\section{Methods}\label{sec4}

\input{sections/methods.tex}

\begin{table*}[p]
\centering
\caption{The 23 model candidates. Source and objective labels follow the frozen experiment registry; published-weight entries identify model families rather than independently verified checkpoint releases. Scratch denotes random initialization.}
\label{tab:model_pool}
\footnotesize
\setlength{\tabcolsep}{4pt}
\begin{tabularx}{\textwidth}{r >{\raggedright\arraybackslash}X l >{\raggedright\arraybackslash}p{2.5cm} >{\raggedright\arraybackslash}p{3.1cm}}
\toprule
ID & Model & Encoder & Objective & Registered source \\
\midrule
745 & OpenMind-T1-MAE & ResEncL & MAE & OpenMind T1 \\
752 & OpenMind-HighIQS-MAE & ResEncL & MAE & OpenMind HighIQS \\
754 & CNN-MAE & ResEncL & MAE & OpenMind Full \\
755 & CNN-MG & ResEncL & Models Genesis & OpenMind Full \\
756 & CNN-S3D & ResEncL & S3D & OpenMind Full \\
757 & CNN-SimCLR & ResEncL & SimCLR & OpenMind Full \\
758 & CNN-SwinUNETR & ResEncL & SwinUNETR SSL & OpenMind Full \\
759 & CNN-VF & ResEncL & Volume Fusion & OpenMind Full \\
760 & CNN-VoCo & ResEncL & VoCo & OpenMind Full \\
761 & ViT-MAE & PrimusM & MAE & OpenMind Full \\
762 & ViT-MG & PrimusM & Models Genesis & OpenMind Full \\
763 & ViT-SimMIM & PrimusM & SimMIM & OpenMind Full \\
764 & ViT-SimCLR & PrimusM & SimCLR & OpenMind Full \\
765 & ViT-SwinUNETR & PrimusM & SwinUNETR SSL & OpenMind Full \\
766 & ViT-VF & PrimusM & Volume Fusion & OpenMind Full \\
767 & ViT-VoCo & PrimusM & VoCo & External VoCo weights \\
768 & BrainIAC & ViT-B/UNETR & BrainIAC & Published checkpoint \\
769 & BrainSegFounder & Swin-T/SwinUNETR & BrainSegFounder & Published checkpoint \\
770 & VoCo-SwinUNETR-L & SwinUNETR-L & VoCo & OpenMind Full \\
771 & SAM-Med3D & ViT-B & SAM-Med3D & Published checkpoint \\
772 & DINO3D-ResEncL-NoGram-E260 & ResEncL-style CNN & DINO (no Gram) & OpenMind epoch 260 \\
773 & ResEncL-Scratch & ResEncL & Scratch & None \\
774 & PrimusM-Scratch & PrimusM & Scratch & None \\
\bottomrule
\end{tabularx}
\end{table*}

\begin{sidewaystable*}[p]
\centering
\caption{The 14 downstream datasets. Source cohort denotes the relevant public release or labelled training partition, whereas study sample denotes the unique participant or paired-case count represented in this benchmark. Image volumes and local manifest entries are shown separately when repeated acquisitions occur. Repeated ISLES and ACDC manifest rows were not counted as independent participants.}
\label{tab:tasks_summary}
\renewcommand{\arraystretch}{0.92}
\fontsize{6.5}{7.1}\selectfont
\setlength{\tabcolsep}{2.5pt}
\begin{tabular}{>{\raggedright\arraybackslash}p{2.0cm} >{\raggedright\arraybackslash}p{2.4cm} >{\raggedright\arraybackslash}p{2.25cm} >{\raggedright\arraybackslash}p{2.1cm} >{\raggedright\arraybackslash}p{2.5cm} >{\raggedright\arraybackslash}p{3.4cm}}
\toprule
Dataset & Benchmark endpoint & Input used & Source cohort & Study sample & Reference target \\
\midrule
\href{https://www.isles-challenge.org/}{ISLES 2022 (ISL)}~\cite{hernandez2022isles} & Stroke-lesion segmentation & DWI and ADC MRI & 250 labelled training cases & 239 unique participants; 239 multichannel volumes; 250 manifest entries & Expert infarct masks \\
\href{https://hntsmrg24.grand-challenge.org/}{HNTS-MRG (HNT)}~\cite{wahid2024hntsmrg} & Primary-tumour and nodal GTV segmentation & Pre-RT and mid-RT T2-weighted MRI & 150 training participants & 150 participants; 300 time-point volumes & GTVp and GTVn masks \\
\href{https://doi.org/10.17632/8bctsm8jz7.1}{MS lesion MRI (MSF)}~\cite{muslim2022ms} & Multiple-sclerosis lesion segmentation & FLAIR MRI & 60 participants & 60 participants; 60 volumes & Consensus manual lesion masks \\
\href{https://topcow24.grand-challenge.org/}{TopCoW (TPC)}~\cite{yang2025benchmarking} & 13-component Circle of Willis segmentation & Paired CTA and MRA & 125 paired training cases & 125 paired participants; 250 volumes & Component-wise vessel masks \\
\href{https://doi.org/10.7937/6BE1-R748}{Yale Brain Metastases (YBM)}~\cite{ramakrishnan2024large} & Brain-metastasis segmentation & T1, post-contrast T1, T2 and FLAIR MRI & 200 participants & 200 participants; 200 four-channel examinations & Brain-metastasis masks \\
Coronary plaque (Dataset602) & Coronary lumen and plaque segmentation & Coronary CTA & Private hospital-curated cohort; 561 patients and 10,936 vessel crops & Development partition: 506 patients and 9,863 crops; source test: 55 patients and 1,073 crops; probe manifest: 100 patients and 200 cases & Lumen, non-calcified-plaque and calcified-plaque masks; quality review ongoing \\
Coronary bridge vessel (Dataset603) & Coronary bridge-vessel segmentation & Coronary CTA & Private hospital-curated cohort; 507 patients and 546 series & 507 patients and 546 series; probe manifest: 100 patients and 108 cases & Binary bridge-vessel masks after merging source instance labels; 44 series from 38 patients lacked matched masks \\
\href{https://sites.wustl.edu/oasisbrains/home/oasis-1/}{OASIS-1 (OAS)}~\cite{marcus2007oasis} & Binary cognitive-status classification & T1-weighted MRI & 416 participants & 235 participants: 135 normal cognition and 100 MCI; probe, 94 & Frozen mapping: 0, normal cognition; 1, MCI \\
\href{https://doi.org/10.7937/TCIA.BDGF-8V37}{UCSF-PDGM (UCS)}~\cite{calabrese2022ucsf} & IDH mutation-status classification & Post-contrast T1 and FLAIR MRI & 495 unique participants; 501 examinations & 414 participants: 330 IDH wild type and 84 IDH mutant; probe, 100 & Histopathological IDH status; frozen mapping: 0, wild type; 1, mutant \\
\href{https://doi.org/10.7937/TCIA.709X-DN49}{UPENN-GBM (UPN)}~\cite{bakas2022upenn} & Overall-survival regression & T1, T1-Gd, T2 and FLAIR MRI & 611 baseline preoperative participants; longitudinal clinical table contains 671 records & 585 baseline scans from deceased participants with numeric survival; probe, 100 & Updated survival from surgery in days; follow-up scans and non-numeric outcomes excluded \\
\href{https://fcon_1000.projects.nitrc.org/indi/retro/sald.html}{SALD (SAL)}~\cite{wei2018sald} & Chronological brain-age regression & T1-weighted MRI & 494 healthy adults & 494 participants; probe, 100 & Age at acquisition in years \\
ATLAS 2.0 (ATL)~\cite{liew2018atlas} & Post-stroke interval regression & T1-weighted MRI & 655 participants with source images; 549 with numeric interval metadata & 210-participant frozen subset; probe, 84; target range, 1--355 days & Release-2.0 field \texttt{Days Post Stroke} \\
\href{https://www.creatis.insa-lyon.fr/Challenge/acdc/}{ACDC (ACD)}~\cite{bernard2018acdc} & Cardiac multi-structure segmentation & End-diastolic and end-systolic cine MRI & 100 labelled training participants; 150 total & 100 participants; 200 unique phase volumes; 400 manifest entries & LV cavity, myocardium and RV cavity masks \\
\href{https://kits19.grand-challenge.org/}{KiTS19 (KIT)}~\cite{heller2021state} & Kidney and kidney-tumour segmentation & Contrast-enhanced abdominal CT & 210 labelled training participants; 300 total & 210 participants; 210 volumes & Kidney and tumour masks \\
\bottomrule
\end{tabular}

\end{sidewaystable*}
\clearpage

\section*{Data availability}
The public datasets are accessible through the providers cited in Table~\ref{tab:tasks_summary}, subject to their access terms. Aggregate results are reported in the text, figures and tables. This preprint does not include the complete model--task score matrix or individual-level data from the two private, hospital-curated coronary CTA cohorts.

% The bibliography database and the required style are included locally.
\bibliography{sn-bibliography}

\end{document}

%% file: sections/main_text.tex
\section{Introduction}\label{sec1}

Self-supervised learning offers a way to develop 3D medical imaging models without requiring expert annotations for every pretraining scan. Such annotations are costly to obtain, particularly for volumetric segmentation, whereas unlabelled scans can support representation learning at a larger scale. By deriving training signals from the images themselves, self-supervised pretraining can learn reusable anatomical and imaging features, reduce reliance on task-specific labels and provide an initialization for downstream adaptation~\cite{xu2025_3dino,tak2026brainiac,zhu2026headct,qiu2026mars}. These advantages have motivated the development of a growing range of foundation models, with different architectures, learning objectives and pretraining collections~\cite{sun2024foundation,hamamci2026ctrate,blankemeier2026merlin,yang2026deciphermr,zhang2025generalist,wald2025openmind}. Their increasing availability creates an important practical question: which pretrained model should be used for a particular medical imaging task?

No single foundation model is consistently best across downstream tasks. Delineating a small lesion, segmenting an organ and predicting a diagnosis or continuous outcome require different aspects of an image representation, so an advantage on one target need not extend to another. Existing evaluation practices make these differences difficult to assess: models are commonly evaluated on different task collections, with different data partitions, preprocessing, fine-tuning settings and comparison baselines~\cite{sun2024foundation,wald2025openmind}. A state-of-the-art result therefore establishes an advantage within the reported setting, but does not establish superiority over models evaluated elsewhere or identify the best model for a new task. Without aligned comparisons, the effects of pretraining are also difficult to distinguish from those of downstream training and evaluation. A broad benchmark with common downstream tasks and clearly specified adaptation protocols is consequently needed to determine which model advantages recur across tasks and to provide a more informative basis for model selection.

Transferability estimation addresses the cost of making a model choice when downstream performance is not yet known. Direct comparison requires fine-tuning each candidate on the target task; repeating this across a large pool can consume substantial training resources. Transferability estimators instead use pretrained outputs or frozen features, often together with target labels, to predict the relative performance expected after adaptation. Established methods summarize source--target label relationships, evidence for simple readouts, class separation or feature information~\cite{tran2019transferability,nguyen2020leep,li2021ranking,you2021logme,bao2019information,pandy2022transferability,huang2022frustratingly,garrido2023rankme}. Segmentation-specific methods such as CC-FV additionally assess feature consistency across cases~\cite{yang2023pick}. Their performance depends on the candidate models, target tasks and evaluation setting~\cite{agostinelli2022stable,chaves2023performance}, motivating a comparison against the same fine-tuned references used to benchmark the foundation models themselves.

The structure of 3D medical tasks creates distinct requirements for transferability estimation. For segmentation, a global feature summary may miss the local distinctions needed to delineate small structures, and abundant background voxels can dominate an unbalanced sample. Spatial and boundary sampling can retain relevant local evidence while limiting computation~\cite{tang2026topology}. For classification and regression, the score must instead assess variation between patients and its relation to the outcome; categorical adaptations of continuous labels can discard outcome ordering. Encoder differences in feature dimension, normalization and spatial resolution also affect the quantities being compared, while covariance estimates can be unstable with limited target samples~\cite{ibrahim2023newer}. These considerations motivate statistics selected for the prediction target and the model comparison. Their value must then be judged both by agreement with the full ranking and by the performance retained among the few models selected for fine-tuning: improved overall ordering does not necessarily improve the top-ranked choices.

Here we establish a benchmark of 23 candidates, comprising 21 pretrained encoders and two scratch controls, across 14 segmentation, classification and regression tasks spanning several anatomies and MRI and CT inputs. Three-fold downstream outcomes provide a common reference for comparisons of architecture, self-supervised objective, data selection and externally pretrained weights under the specified adaptation protocols. No candidate ranks among the top three on every task. Matched comparisons use OpenMind, a large collection of 3D head-and-neck images for self-supervised pretraining, as a common data source~\cite{wald2025openmind}. They contrast ResEncL, a large residual-encoder U-Net representing 3D convolutional neural networks (CNNs), with Primus-M, a medium-sized 3D vision transformer~\cite{isensee2024nnu,wald2025primus}. ResEncL shows a broad performance advantage in these comparisons, while masked-image-pretrained Primus-M models perform better on segmentation than their counterparts trained with other objectives. Pretraining-data selection also changes which downstream tasks benefit most. The benchmark therefore provides evidence for both model development and task-specific model choice.

We also benchmark established transferability estimators and introduce InsightRank, which combines target-appropriate frozen-feature statistics through architecture-aware pairwise comparisons. Segmentation uses local voxel evidence, whereas classification and regression use patient-feature variation and neighbour-outcome agreement. Evaluation covers full-ranking agreement, three-model shortlist performance and recorded computational cost. InsightRank retains strong segmentation models with low shortlist regret and reduces recorded extraction-plus-training time in the measured segmentation workflow. For classification and regression, it improves mean ranking agreement over the matched conventional estimators but does not recover a task winner in the three-model shortlists. Together, these analyses establish a common basis for comparing foundation models and for assessing how well transferability estimation supports a limited fine-tuning budget.

\section{Results}\label{sec2}
\begin{figure*}[!tp]
\centering
\includegraphics[width=\textwidth]{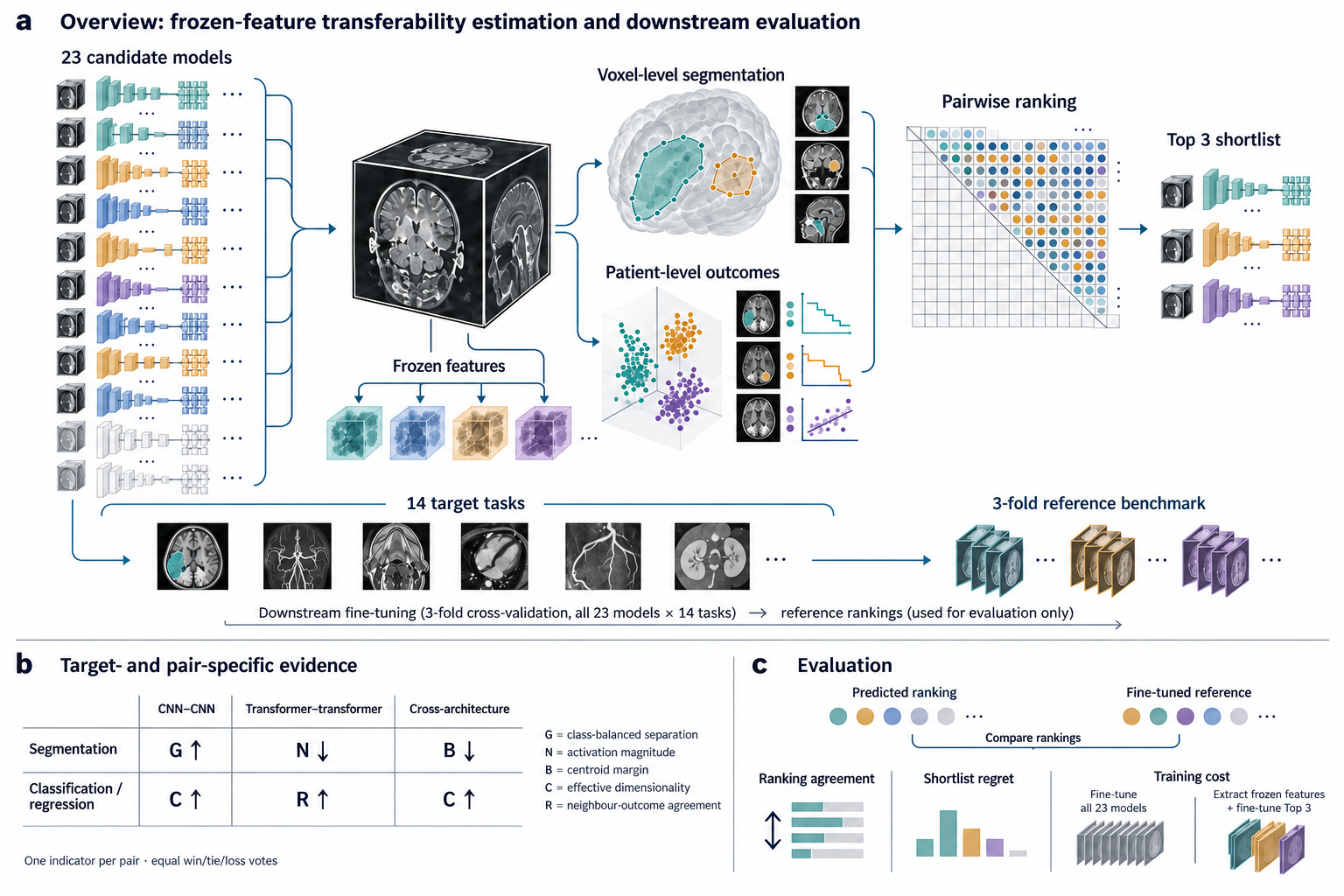}
\caption{\textbf{Study overview: benchmarking and transferability estimation for 3D medical imaging.} \textbf{a}, Two complementary workflows use the same pool of 21 pretrained encoders and two scratch controls. Frozen features provide voxel-level evidence for segmentation and patient-level evidence for classification and regression, from which InsightRank produces pairwise rankings and a three-model shortlist. Separately, downstream fine-tuning of all 23 candidates across 14 tasks yields 322 three-fold mean outcomes and the reference rankings used to evaluate model selection. The reference benchmark covers the full candidate pool, not only shortlisted models. \textbf{b}, Indicator assignments depend on the target and encoder pair: segmentation uses class-balanced nearest-centroid accuracy ($G$), activation amplitude ($N$) and centroid margin ($B$); classification and regression use effective dimensionality ($C$) and neighbour-outcome agreement ($R$). Arrows indicate the preferred direction. Pairwise wins, ties and losses are aggregated with equal weight. \textbf{c}, Evaluation compares predicted and fine-tuned rankings, measures performance loss at a fixed shortlist size, and compares recorded extraction-plus-training time with exhaustive fine-tuning where timing data are available. Images, outcome curves, matrix colours, example shortlists and charts are schematic; they do not show measured predictions, survival-curve estimates or numerical results. Task-specific evaluation and timing coverage are detailed in Methods.}
\label{fig:benchmark_landscape}
\end{figure*}

We evaluated 23 model candidates across 14 downstream tasks under a shared benchmark, treating voxel-level segmentation and image-level classification or regression as distinct prediction settings (Fig.~\ref{fig:benchmark_landscape}a). No single foundation model ranked among the top three on every task; controlled contrasts within the pool also revealed recurring differences associated with architecture, pretraining objective and data selection. We then used the fine-tuned rankings to benchmark established transferability estimators and to evaluate InsightRank, which ranks candidates from target-appropriate frozen features before fine-tuning. Finally, we evaluated the quality of three-model shortlists and compared their recorded training burden with exhaustive fine-tuning.

\subsection{A controlled benchmark of 23 models across 14 medical imaging tasks}
\begin{figure*}[!t]
\centering
\includegraphics[width=\textwidth]{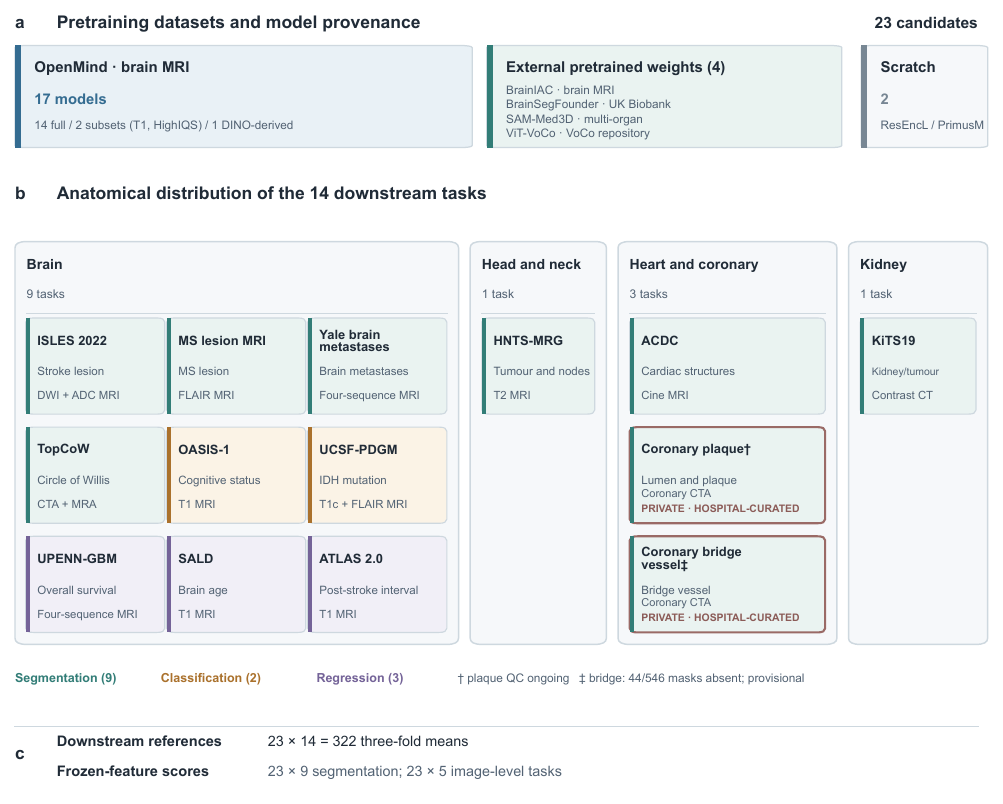}
\caption{\textbf{Pretraining provenance and downstream task atlas.} \textbf{a}, The registered model pool contains 17 OpenMind-derived checkpoints (14 trained with the full collection, two using T1-weighted or image-quality-filtered subsets, and one DINO-derived), four external pretrained-weight models and two randomly initialized controls. ViT-VoCo inherits weights from the VoCo GitHub repository rather than OpenMind; the exact upstream checkpoint variant used here was not independently verified. Counts denote model candidates, not pretraining images. \textbf{b}, The 14 downstream datasets are grouped by anatomical target: nine brain, one head-and-neck, three heart or coronary, and one kidney task. TopCoW is assigned to the brain group because it segments the intracranial Circle of Willis. Card colour denotes segmentation (teal), classification (ochre) or regression (violet); each card identifies its target and imaging input. The coronary-plaque and coronary bridge-vessel CTA datasets are private, hospital-curated in-house datasets, as indicated on their cards. Plaque annotations remain under quality review (\textdagger). For the provisional bridge-vessel task (\textdaggerdbl), 44 of 546 series lacked matched masks and were encoded as all-background labels in training and evaluation rather than excluded. \textbf{c}, All 23 candidates have three-fold downstream references on all 14 tasks (322 model--task means), and frozen-feature scoring covers the same candidates on nine segmentation and five image-level tasks.}
\label{fig:study_atlas}
\end{figure*}

We designed the benchmark to separate several sources of variation in model suitability: encoder architecture, self-supervised objective and pretraining-data selection on one side, and anatomy, acquisition and prediction target on the other (Fig.~\ref{fig:study_atlas}). All 23 candidates were evaluated on the same 14 downstream tasks, yielding 322 three-fold model--task means. Within each task, candidates are compared using its designated metric---Dice for segmentation, AUROC for classification and mean absolute error for regression---so that changes in rank can be examined across tasks with different outcome scales.

The model pool was assembled around interpretable contrasts, rather than a single leaderboard (Table~\ref{tab:model_pool}). It includes 21 pretrained encoders and two architecture-matched scratch controls: 11 ResEncL-like convolutional networks and 12 transformers. ResEncL is the large residual-encoder U-Net, a 3D convolutional network with a hierarchy of spatial resolutions; Primus-M is a medium-sized, transformer-based 3D image encoder that processes high-resolution image tokens with attention~\cite{isensee2024nnu,wald2025primus}. OpenMind provides a common pretraining source, with approximately 114,000 three-dimensional head-and-neck volumes from OpenNeuro studies spanning T1-weighted, T2-weighted, FLAIR, diffusion-derived and angiographic images~\cite{wald2025openmind}. Among 14 full-OpenMind checkpoints, five ResEncL--PrimusM pairs share the registered data source and self-supervised objective, allowing architecture-associated differences to be examined within each objective. Multiple objectives within each architecture family provide a complementary comparison of pretraining methods. Three ResEncL MAE checkpoints trained on the full collection, a T1-weighted subset or an image-quality-filtered subset probe how data composition changes transfer under a common architecture and objective. The remaining OpenMind-derived DINO3D checkpoint, four externally pretrained models and two scratch controls broaden the comparison to distinct pretraining sources and initialization choices. The external models are BrainIAC, BrainSegFounder, SAM-Med3D and ViT-VoCo; the latter inherits VoCo repository weights rather than OpenMind weights~\cite{tak2026brainiac,cox2024brainsegfounder,wang2023sammed3d,wu2024voco}.

The task pool was selected to test whether those model contrasts persist as the information required for prediction changes (Fig.~\ref{fig:study_atlas}b; Table~\ref{tab:tasks_summary}). Its nine segmentation tasks range from focal lesions and tumour contours to cardiac structures and narrow arterial trees, varying the spatial extent and boundary precision of the target. Two classification and three regression tasks instead assign an outcome to an entire image, including disease status, brain age, post-stroke interval and survival. Nine tasks involve the brain or intracranial vessels, permitting comparisons among different targets within a predominantly brain-imaging domain; head-and-neck, cardiac, coronary and kidney tasks probe transfer beyond that domain. MRI, CT and paired CTA/MRA inputs introduce acquisition differences, although anatomy and modality also change together in some cross-domain comparisons. The two coronary CTA datasets are private, hospital-curated cohorts. Together, these axes allow us to examine whether a model favored for one anatomical structure or prediction scale remains suitable for another, without treating every cross-task contrast as an isolated effect of modality or organ.

\subsection{No foundation model ranks in the top three on every downstream task}
No candidate ranked among the top three across all 14 downstream tasks (Fig.~\ref{fig:motivation}a). Rankings changed within both voxel-level segmentation and image-level classification or regression, as well as between these task groups. A model that performed strongly on one task therefore could not be assumed to perform equally well on another.

Among the nine segmentation tasks, seven different candidates ranked first, and no fixed model entered the Top~3 on more than four tasks (Fig.~\ref{fig:motivation}b). The winners differed even among brain MRI datasets: CNN-SimCLR led MS lesion segmentation, VoCo-SwinUNETR-L led ischaemic-stroke lesion segmentation, and CNN-MG led brain-metastasis segmentation. Individual models also changed rank sharply across spatial targets. CNN-SimCLR ranked first for MS lesions but 21st for the intracranial arterial tree, whereas ViT-SimMIM ranked first for the arterial tree and kidney/tumour masks but 16th for MS lesions. A strong result for one type of mask therefore gave little assurance of a similarly strong result for another, whether the target was a lesion, a vessel or an organ.

Classification and regression likewise had different leaders across their five tasks (Fig.~\ref{fig:motivation}c). DINO3D led MCI classification, brain-age estimation and post-stroke interval estimation; CNN-SwinUNETR led IDH classification, and CNN-MG led overall-survival regression. DINO3D ranked second for IDH but only 14th for survival, despite leading the other two regression tasks. The most relevant distinction is not classification versus regression: brain age and time since stroke can be reflected in features present at imaging, whereas survival also depends on events and treatment after the baseline scan~\cite{wen2024brainage,bernstock2025gabapentinoids}. Within image-level prediction, the relation between the scan and the clinical outcome is therefore important to model choice.

The scratch controls show that task dependence also applies to the value of pretraining. ResEncL-Scratch placed second for brain age and third for post-stroke interval, ahead of all seven full-OpenMind-pretrained ResEncL checkpoints on both tasks; PrimusM-Scratch placed second for survival, ahead of all full-OpenMind PrimusM checkpoints. Yet ResEncL-Scratch ranked near the bottom for kidney/tumour and intracranial-vessel segmentation. These controls do not imply that training from scratch is generally preferable. They show that pretraining cannot be assumed to improve every downstream target, reinforcing the need to compare candidate initializations on the task of interest.

\subsection{Architecture and pretraining choices distinguish model performance}
\begin{figure*}[!tp]
\centering
\includegraphics[width=\textwidth]{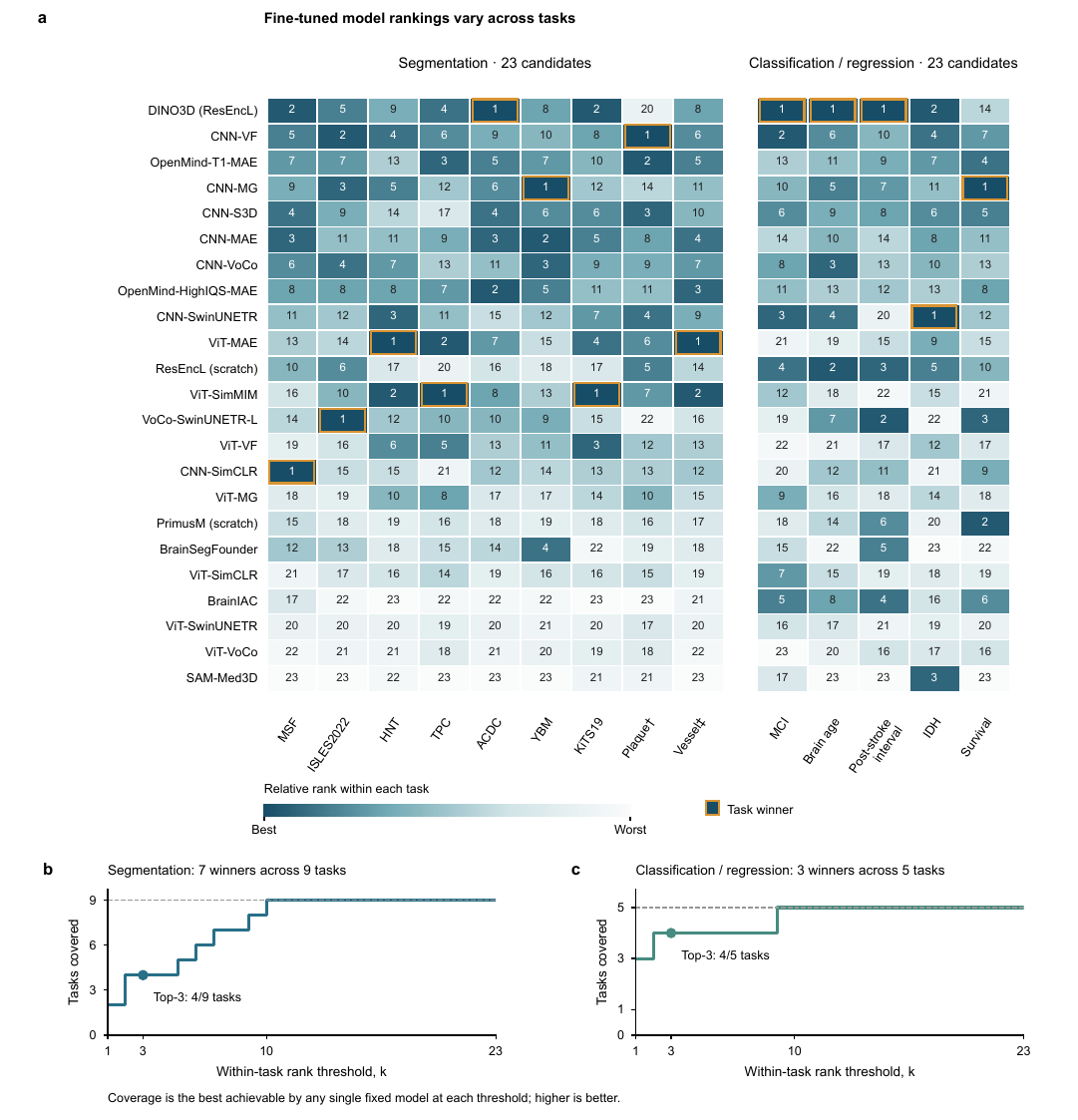}
\caption{\textbf{Cross-task benchmark of 3D medical model candidates.} \textbf{a}, Ranks of three-fold mean downstream outcomes for 23 candidates across nine segmentation and five classification or regression tasks. Numbers show within-task ranks; colour shows $(r-1)/(M-1)$, with rank $r$ and candidate count $M=23$. Outlines identify winners. Scratch models are non-pretrained controls. \textbf{b,c}, Maximum task coverage of any single fixed model at rank threshold $k$. The maximizing model may change with $k$, so each curve is an optimistic envelope rather than one model's performance. Dagger and double dagger mark the coronary annotation qualifications detailed in Methods.}
\label{fig:motivation}
\end{figure*}

The model pool reveals several recurring patterns across the 14 tasks (Fig.~\ref{fig:motivation}a). We examined architecture within matched OpenMind pretraining objectives, objectives within each encoder family, subsets of OpenMind data under ResEncL MAE, and externally released weights under the common downstream protocol. We then compared how consistently individual models ranked well across tasks. These views distinguish broad model-family advantages from the pretraining and task choices that modify them.

ResEncL had a substantial overall advantage over Primus-M in this benchmark. Across five pairs pretrained on full OpenMind with the same self-supervised objective, the ResEncL model ranked higher in 57 of 70 task-wise comparisons, including 35 of 45 segmentation and 22 of 25 image-level comparisons. The advantage therefore extended beyond one objective, organ or prediction scale. ResEncL's residual convolutional stages retain feature maps at several resolutions, from local tissue patterns to broader anatomical context. This gives dense decoders access to fine spatial detail while deeper features can summarize larger structures for image-level prediction. Primus-M instead processes image tokens through attention and reconstructs dense outputs from its token representation~\cite{wald2025openmind,wald2025primus}. These design differences are consistent with the CNN's advantage across both prediction scales, although the comparison does not separate encoder design from decoder and optimization choices. MAE narrowed or reversed the advantage on some segmentation targets, but did not overturn the overall pattern.

Pretraining objective mattered most visibly for Primus-M segmentation. With the encoder family and full OpenMind source held fixed, masked-image objectives SimMIM and MAE had mean ranks of 6.7 and 7.0 across the nine segmentation tasks, compared with 17.0 for SimCLR and 19.7 for the SwinUNETR objective. Reconstructing missing image regions requires the token representation to retain spatial detail, which is consistent with their stronger dense-prediction results~\cite{wald2025openmind}. ResEncL showed a different pattern: Volume Fusion ranked well in both segmentation and image-level prediction, with mean ranks of 5.7 and 5.8. Thus objective choice was important, but its most consistent effect was tied to the encoder and the spatial demands of the downstream task.

Pretraining-data selection produced a shared pattern across the two smaller ResEncL MAE subsets. Both the T1-weighted and image-quality-filtered checkpoints outranked the full-data MAE checkpoint on MCI, post-stroke interval and survival prediction, as well as on stroke-lesion and intracranial-artery segmentation. Both ranked below it on brain-age estimation, MS lesions, brain metastases and kidney/tumour segmentation. These repeated gains and losses show that curating the pretraining corpus changed the downstream task profile rather than yielding a uniform improvement. In particular, the T1-only checkpoint did not improve T1-based brain-age prediction, so simple sequence matching did not account for the observed pattern.

External pretrained weights showed a disadvantage as a group on dense prediction under this benchmark's downstream protocol. The four external candidates had a mean rank of 19.8 across segmentation tasks, compared with 15.4 across classification and regression. BrainIAC illustrates the task-type difference within one checkpoint, with mean ranks of 21.7 and 7.8, respectively. OpenMind-derived candidates led all 14 tasks, including head-and-neck, cardiac and kidney segmentation outside the predominantly brain-imaging pretraining domain. This compares complete released configurations, not pretraining data alone: the external checkpoints also differ in architecture, objective and processing pipeline. ViT-VoCo inherits external VoCo weights rather than OpenMind weights.

Finally, frequent high rankings and consistency across tasks captured different aspects of transfer. Across all 14 tasks, DINO3D entered the Top~3 seven times and had a mean rank of 5.6, but ranked 14th for survival. CNN-VF entered the Top~3 three times and had a similar mean rank of 5.7, yet never fell below 10th. These similar averages conceal different profiles: DINO3D achieved more high placements, whereas CNN-VF avoided such a sharp drop on any task. Reporting both the mean and the spread of task-wise ranks helps distinguish a high-performing but uneven candidate from a steadier default when task-specific fine-tuning results are not yet available.

\subsection{Benchmarking established transferability estimators on 3D medical tasks}
The benchmark establishes how each model performs after fine-tuning, but obtaining this ranking for a new task would require training the full candidate pool again. Transferability estimation addresses this selection problem by using a small labelled target set and frozen model features to predict which pretrained models are worth fine-tuning. We tested whether established estimators recover the observed three-fold downstream rankings of the same 23 candidates. We analysed segmentation separately from classification and regression because a useful segmentation representation must support voxel-level distinctions, whereas the other tasks assign one outcome to each image. This comparison asks not only whether an estimator reproduces the full ranking, but also whether its highest-ranked models include a strong choice for the target task.

The comparators summarize different properties of frozen features: evidence for a simple readout, separation or overlap of labelled groups, and consistency of feature distributions across cases~\cite{you2021logme,bao2019information,pandy2022transferability,li2021ranking,yang2023pick}. We adapted the applicable scores to the available 3D representations, including LEEP-family variants, rather than assuming that a self-supervised encoder supplies source-class predictions. The common difficulty is that a favourable global feature statistic need not indicate whether neighbouring voxels preserve a small structure or its boundary. Moreover, a full 3D volume contains many correlated voxels and often far more background than foreground; a score computed without spatially and class-aware sampling can be dominated by the latter or become impractical to calculate. These concerns are most acute for segmentation~\cite{tang2026topology3d}. For classification and regression, spatial boundaries are not the immediate target, but a score based on categorical group separation can still miss the ordering of continuous outcomes. The following results therefore test each comparator against the task type for which it was computed, rather than treating its definition as evidence of transferability.

For segmentation, we applied CC-FV, LogME, GBC and H-score to shared class-stratified patient probes and a final-layer representation (Methods). On the predefined four-task comparison, mean Kendall $\tau$-b was $-0.107$ for LogME, $-0.188$ for GBC and $-0.097$ for H-score. CC-FV reached 0.447, but its three-model shortlist incurred 1.253 Dice percentage points of mean regret (Table~\ref{tab:segmentation_results}). Thus a method designed for segmentation recovered more of the overall ranking than the other conventional scores on this panel, without reliably retaining the best candidate at the three-model budget. The CC-FV comparator here uses a diagonal-Gaussian approximation, rather than an exact reproduction of the published full-covariance formulation.

For classification and regression, each patient contributed one feature vector and one outcome. We evaluated LogME, H-score, GBC and NLEEP, retaining continuous regression targets for LogME but converting them to outcome quintiles for the three categorical comparators (Methods). Across five tasks, deepest-layer LogME achieved mean Kendall $\tau$-b of 0.055; averaging four layers raised it to 0.105. H-score, GBC and NLEEP yielded $-0.025$, $-0.189$ and $-0.037$, respectively (Table~\ref{tab:imagelevel_23_pool}). Unlike dense segmentation, these tasks do not require a score to preserve voxel adjacency, but the readout and distributional assumptions still yielded weak agreement with the observed model orderings. For regression, outcome binning additionally discards within-bin ordering. Together, the two task panels motivate evaluating target-scale evidence rather than transferring one feature summary unchanged across prediction types.

\subsection{InsightRank combines target- and architecture-aware evidence}
InsightRank compares frozen representations at the scale of the downstream target, with the same eligible patients supplying features for every candidate (Fig.~\ref{fig:benchmark_landscape}a,b). Segmentation uses sampled voxel features and shared ground-truth locations; classification and regression use one vector per patient. The ranking rule assigns an indicator by target type and encoder-pair type, then combines pairwise wins, ties and losses without fitted fusion weights. This design avoids forcing voxel-level and patient-level evidence onto one numerical scale, while retaining a common ranking and three-model shortlist for both task groups.

For segmentation, $G$ measures class-balanced nearest-centroid accuracy, $B$ measures the margin to the nearest competing class, and $N$ measures activation amplitude under a fixed extraction protocol. Shared spatial coordinates and boundary samples retain local label evidence while limiting computation. Within convolutional pairs, the rule uses $G$; within transformer pairs it uses $-N$; and across architectures it uses $-B$. These assignments were fixed during development, not derived uniquely from the definitions of the indicators (Methods). They allow the relevant frozen-feature property to differ by comparison rather than imposing one voxel statistic on all pairs.

\begin{figure*}[t]
\centering
\includegraphics[width=\textwidth]{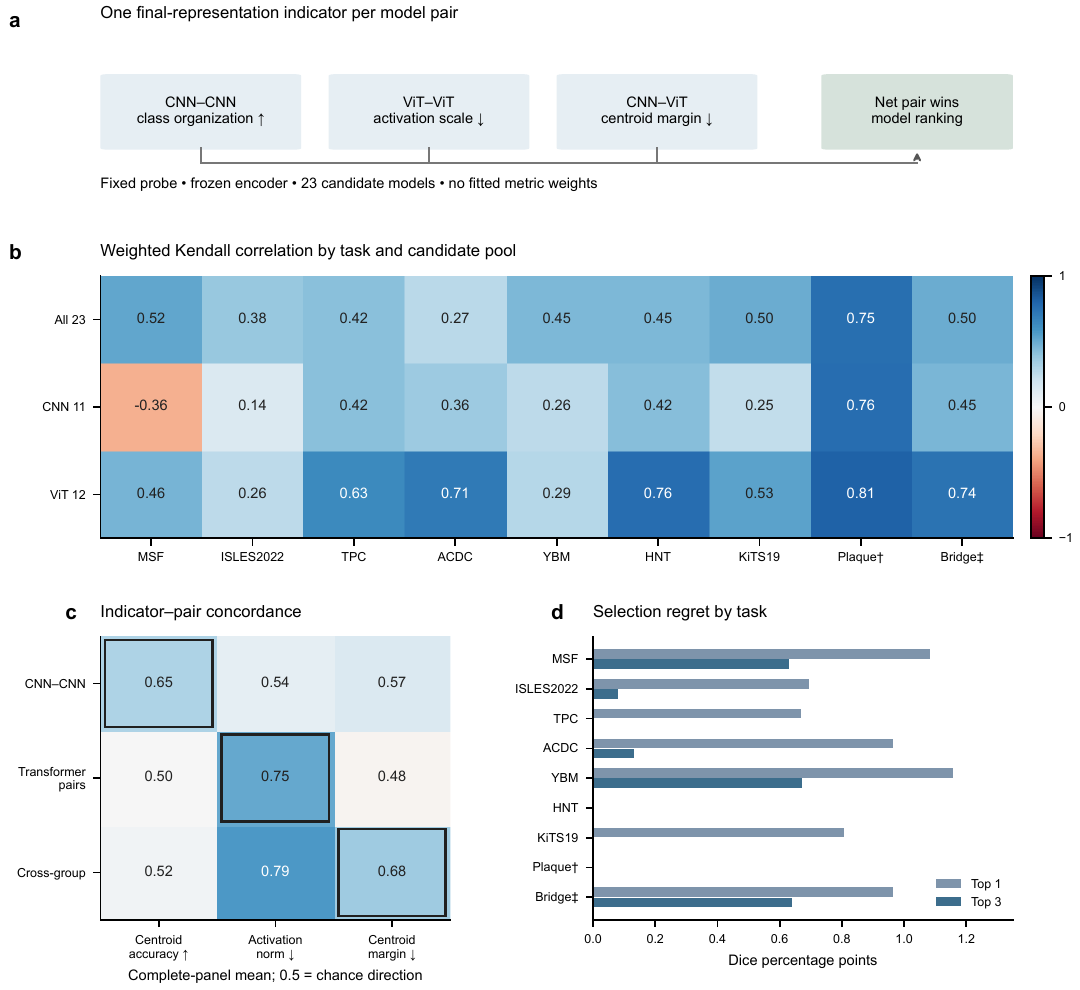}
\caption{\textbf{Comparison-dependent evidence for spatial targets.} \textbf{a}, Three feature statistics assigned to model-pair types, followed by the shared net-win aggregation. \textbf{b}, Weighted Kendall correlations for the complete pool and within-architecture rankings across nine tasks. \textbf{c}, Pairwise concordance of each statistic, averaged across tasks; outlines mark fixed assignments. Predicted ties count as half agreement and 0.5 denotes chance-direction agreement. \textbf{d}, Top~1 and Top~3 regret in Dice percentage points. All candidates use the same task-specific probe and three-fold reference. Both coronary annotation qualifications are detailed in Methods.}
\label{fig:segmentation_ranking}
\end{figure*}

For segmentation, InsightRank tracked rankings obtained after fine-tuning. Across nine tasks, mean Kendall $\tau$-b was 0.431, Spearman correlation was 0.575 and weighted Kendall correlation was 0.470 against three-fold downstream references (Fig.~\ref{fig:segmentation_ranking}). Excluding the two scratch controls retained mean weighted correlation of 0.488, so agreement was not confined to distinguishing pretrained from randomly initialized models. On the predefined four-task comparison, its weighted Kendall correlation was 0.549, compared with 0.365 for CC-FV and negative values for LogME, GBC and H-score (Table~\ref{tab:segmentation_results}). The comparison is specific to this panel; the complete nine-task analysis and four-task comparator analysis are not interchangeable.

Encoder-pair design changed which spatial feature property aligned with downstream ordering. Within convolutional encoders, class-balanced nearest-centroid accuracy achieved mean weighted correlation of 0.302; within transformers, negative activation norm achieved 0.578. The assigned indicators agreed with downstream pair ordering in 65\% of convolutional pairs, 75\% of transformer pairs and 68\% of cross-architecture pairs. These associations support comparison-dependent evidence, without showing that any assigned statistic was uniquely optimal.

For classification and regression, InsightRank uses patient-level rather than voxel-level evidence. Effective dimensionality ($C$) summarizes the breadth of variation across patient features without using outcomes; neighbour-outcome agreement ($R$) tests whether patients near one another in feature space have similar labels or continuous outcomes. The fixed rule uses $R$ for transformer--transformer pairs and $C$ for other pairs, so neither statistic is treated as universally sufficient. Across all 23 candidates and five tasks, mean Kendall $\tau$-b was 0.352, Spearman correlation was 0.465 and weighted Kendall correlation was 0.310 (Table~\ref{tab:imagelevel_23_pool}). All three mean correlations exceeded those of each matched conventional estimator, including the four-layer LogME variant. This is an improvement in full-ranking agreement, not evidence that the highest-ranked image-level candidates always include the task winner.

\begin{table*}[t]
\centering
\caption{Image-level transferability results for the full 23-candidate benchmark across five classification and regression tasks. Correlations and regret are task-equal means. Top~3 hit counts tasks in which the predicted shortlist contains a best fine-tuned model; Top~1 regret is normalized by the observed performance range within each task. All methods use the same candidate pool. These tasks contributed to rule development, so the results are not independent task validation; values are descriptive and no hypothesis tests were performed.}
\label{tab:imagelevel_23_pool}
\scriptsize\setlength{\tabcolsep}{4pt}
\begin{tabularx}{\textwidth}{@{}Xrrrrr@{}}
\toprule
Estimator & \shortstack{Kendall\\$\tau$-b} & \shortstack{Spearman\\$\rho$} & \shortstack{Weighted\\Kendall $\tau$} & \shortstack{Top~3\\hit / 5} & \shortstack{Top~1\\regret} \\
\midrule
InsightRank & 0.3521 & 0.4654 & 0.3099 & 0/5 & 0.3118 \\
LogME, deepest layer & 0.0545 & 0.0700 & 0.0167 & 0/5 & 0.4834 \\
LogME, four-layer mean & 0.1051 & 0.1202 & 0.0700 & 0/5 & 0.3999 \\
H-score, deepest layer & -0.0245 & -0.0128 & -0.0582 & 1/5 & 0.4956 \\
GBC, deepest layer & -0.1889 & -0.2994 & -0.2044 & 0/5 & 0.6320 \\
NLEEP, deepest layer & -0.0372 & -0.0429 & -0.0143 & 0/5 & 0.4424 \\
\bottomrule
\end{tabularx}
\end{table*}

\subsection{Three-model shortlists retain strong segmentation models but miss image-level winners}
Rank correlation measures the ordering of all candidates, but the practical decision is which few models to train. We therefore evaluated the best downstream result retained by a three-model shortlist, separately for segmentation and image-level targets (Fig.~\ref{fig:benchmark_landscape}c).

\begin{figure*}[t]
\centering
\includegraphics[width=\textwidth]{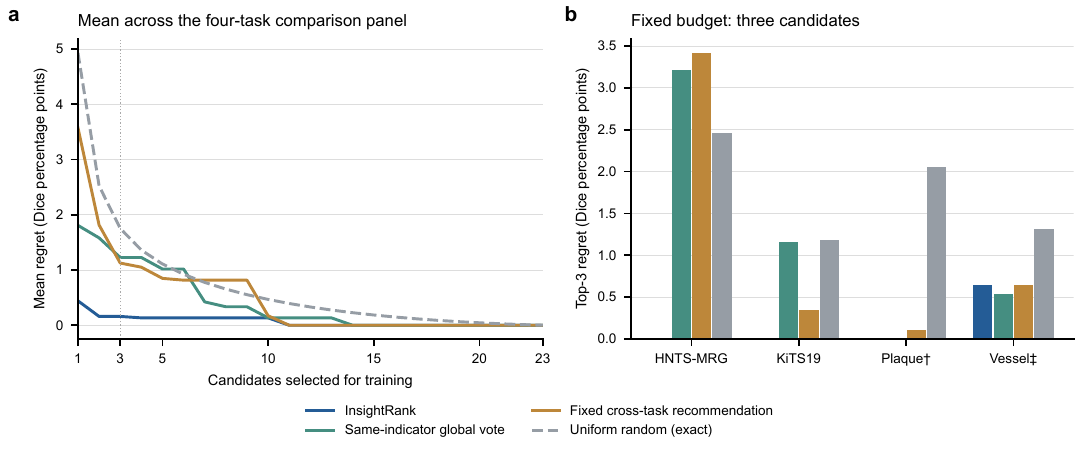}
\caption{\textbf{Selection loss at a matched training budget.} \textbf{a}, Mean regret versus shortlist size in the predefined four-task comparison. The random curve is an exact expectation over uniformly sampled subsets. The fixed recommendation uses ranks from other tasks only, as specified in Methods. \textbf{b}, Task-level Top~3 regret. All methods use the same downstream reference; no additional models were trained. Both coronary tasks retain their annotation qualifications.}
\label{fig:budget}
\end{figure*}

For segmentation, InsightRank's shortlist contained the best of 23 models on four of nine tasks. Its best selected model fell a mean of 0.239 Dice percentage points below the task winner across the nine tasks. In the predefined four-task comparison, pair-specific indicator assignment reduced mean Top~3 regret to 0.160 points, from 1.227 when the same indicators voted globally (Table~\ref{tab:segmentation_results}). At the same budget, CC-FV incurred 1.253 points, random selection 1.753 and a fixed cross-task recommendation 1.126 (Fig.~\ref{fig:budget}). Both non-coronary tasks in this panel had zero regret for InsightRank. Weighted correlation improved from 0.434 with global voting to 0.549, although Spearman correlation decreased from 0.654 to 0.562; this contrast shows why rank agreement alone is not a sufficient selection endpoint.

\begin{table*}[t]
\caption{Controlled model selection in the predefined four-task spatial-target comparison. All methods rank the same 23 candidates. Correlations are task means; regret is in Dice percentage points. Two tasks retain annotation qualifications. Task identities, evaluation roles and comparator implementations are specified in Methods.}
\label{tab:segmentation_results}
\centering\scriptsize\setlength{\tabcolsep}{4pt}
\begin{tabularx}{\textwidth}{@{}Xrrrrrr@{}}
\toprule
Method & \shortstack{Weighted\\$\tau$} & $\tau$-b & $\rho$ & \shortstack{Top-3\\hit} & \shortstack{Top-1\\regret} & \shortstack{Top-3\\regret} \\
\midrule
\textbf{InsightRank} & 0.5492 & 0.4328 & 0.5623 & 3/4 & 0.443 & 0.160 \\
\input{analysis/segmentation/segmentation_comparator_rows.tex}
CC-FV, diagonal & 0.3654 & 0.4466 & 0.6040 & 0/4 & 2.165 & 1.253 \\
LogME & -0.0629 & -0.1067 & -0.1487 & 1/4 & 4.997 & 1.526 \\
GBC, diagonal & -0.0522 & -0.1877 & -0.2720 & 2/4 & 3.745 & 1.121 \\
H-score & -0.0591 & -0.0968 & -0.1354 & 1/4 & 4.997 & 1.526 \\
\bottomrule
\end{tabularx}
\end{table*}

For classification and regression, the shortlist result was weaker. Although InsightRank had higher mean full-ranking correlations than the matched conventional estimators, its Top~3 did not contain the best fine-tuned model on any of the five tasks (Table~\ref{tab:imagelevel_23_pool}). H-score contained a winner on one task. Top~3 regret was not available for the 23-model image-level panel, so the magnitude of the missed performance cannot be compared with the Dice-point losses above. Improved ordering did not, in this panel, establish a useful three-model choice.

\subsection{Fine-tuning three selected models reduces recorded segmentation job time}
Selecting three candidates using InsightRank avoids fine-tuning the entire 23-model pool, but requires extracting frozen features from every candidate. We compared the recorded job time of these two workflows where complete logs were available. Fine-tuning three rather than 23 models reduces the number of candidate-training runs by 87\%; the time comparison below also includes feature extraction for all 23 candidates.

For segmentation, exhaustive adaptation across seven tasks accumulated 1,820.1 hours of recorded job time. Training the selected models took 245.5 hours and extracting features from all candidates took 12.4 hours, giving a 257.8-hour subtotal---85.8\% below exhaustive adaptation for the workflows illustrated in Fig.~\ref{fig:benchmark_landscape}c. These are summed job durations rather than elapsed wall-clock time; feature-statistic computation was not timed and is excluded. The reduction thus describes the recorded seven-task workflow, not a prospective end-to-end speedup for all 14 tasks.

For classification and regression, a matched full-pool end-to-end timing comparison was not available. We therefore do not extrapolate the segmentation job-time reduction to those five tasks.

\section{Discussion}\label{sec3}
Evaluating foundation models on common downstream tasks reveals both recurring advantages in model design and limits to their applicability across targets. Our benchmark addresses the lack of directly comparable evidence across independently reported evaluations by placing pretrained encoders and scratch controls against the same set of segmentation, classification and regression targets. No candidate ranked among the top three on every task, yet the matched comparisons identified consistent advantages associated with architecture and pretraining objective. These findings make the benchmark useful both for developing models and for choosing among existing ones. The accompanying evaluation of transferability estimators asks whether that choice can be informed before undertaking full fine-tuning. InsightRank improved several ranking and selection outcomes, with stronger evidence for practical three-model selection in segmentation than in classification and regression.

Architecture, self-supervised objective and pretraining-data composition should be considered jointly when designing 3D medical foundation models. ResEncL outperformed Primus-M in 57 of 70 matched-objective task comparisons, indicating a broad advantage under the evaluated training protocols. Its hierarchy of convolutional feature maps offers a plausible explanation: local tissue detail and broader anatomical context remain available at different resolutions for downstream prediction. Within Primus-M, the stronger segmentation results of MAE and SimMIM are consistent with reconstruction objectives encouraging the retention of spatial information~\cite{wald2025openmind,wald2025primus}. Together, these observations suggest that the benefit of an objective depends on the architecture through which it is learned. Data selection further modifies this relationship. The T1-weighted and quality-filtered ResEncL MAE checkpoints shared gains on several targets and losses on others relative to full-data pretraining; neither subset yielded a uniform improvement. Similarly, external weights performed less well as a group on segmentation than on classification and regression under the common downstream protocol. That comparison concerns complete released configurations, including their architectures and processing requirements. The practical implication is to assess a pretraining strategy across a representative range of targets, rather than use corpus size or modality matching alone as a guide to expected transfer.

Choosing a downstream model also requires distinguishing broad reliability from strength on particular targets. In segmentation, lesions, organs and arterial trees place different demands on local contrast, spatial extent and boundary precision; the different leaders within brain MRI show that a shared organ and modality do not ensure a shared best initialization. For classification and regression, the connection between the scan and the endpoint is equally relevant. DINO3D's strong performance on brain age and post-stroke interval, followed by its lower position on survival, is consistent with a difference between properties reflected at imaging and outcomes also shaped by subsequent events and treatment~\cite{wen2024brainage,bernstock2025gabapentinoids}. Across tasks, DINO3D and CNN-VF had similar mean ranks but different performance distributions: the former achieved more top-three placements, whereas the latter avoided large drops in rank. Reporting both average and worst-task performance therefore helps distinguish a broadly reliable starting point from a model with greater strengths on selected targets. The competitive scratch controls on several regression tasks reinforce the need to establish the benefit of pretraining for the intended endpoint.

The transferability benchmark indicates that useful model-selection evidence must reflect both the prediction target and the architectures being compared. For segmentation, scores based on global feature distributions can overlook the local distinctions needed to delineate small structures and boundaries. InsightRank samples shared voxel locations and boundaries, then selects the feature statistic according to the encoder pair; the improvement over global voting of the same indicators supports the value of this comparison rule. For classification and regression, the relevant evidence instead concerns variation between patients and the association of that variation with outcomes. Combining effective dimensionality with neighbour-outcome agreement improved mean ranking correlations over the matched conventional estimators. These findings support using different properties of frozen representations for different comparisons. They do not require the properties to share a numerical scale, because aggregation uses pairwise orderings. The assignments remain empirically determined, making their transfer to independent tasks a central question for further evaluation.

The practical value of transferability estimation depends on the downstream performance retained at a given training budget. For segmentation, InsightRank's three-model shortlist had a mean loss of 0.239 Dice percentage points relative to exhaustive selection across nine tasks. In the seven-task timing subset, extracting features from all candidates and fine-tuning the selected three reduced recorded job time by 85.8\%. These results show how a small candidate budget can retain strong segmentation performance while reducing training work. The time reduction covers recorded extraction and training durations, with statistic computation untimed. Classification and regression expose a different aspect of the selection problem: higher whole-ranking agreement did not place a task winner in any of the five three-model shortlists. Because Top~3 regret was unavailable for these tasks, the performance cost of missing those winners remains unresolved. Ranking correlation, shortlist performance loss and computational cost should therefore be evaluated together. Their joint reporting distinguishes an estimator that orders candidates reasonably well from one that reliably directs a limited fine-tuning budget toward strong models.

The main limits to generalization concern benchmark coverage, reference quality and independence of method validation. The candidate pool is centred on OpenMind-derived models, and the task set remains weighted toward brain imaging. Matched comparisons reduce variation in selected factors, but decoder, optimization and data differences prevent causal attribution to a single design choice. Three-fold means capture variation across the evaluated partitions, without measuring repeated-training uncertainty, and some imported references lacked source files for independent split and checkpoint verification. The coronary results also require specific qualification: plaque annotations remain under review, and 44 bridge-vessel series without matched masks were encoded as all-background during training and evaluation, leaving that task provisional. For InsightRank, the four segmentation evaluation tasks were held apart from rule development, whereas all five classification and regression tasks contributed to development. Extraction conventions and the architecture composition of the candidate pool can also affect its scores. Further assessment should freeze the rule before testing independent tasks and model families, verify reference data and annotations, and measure the complete selection-to-fine-tuning workflow.

Common downstream evaluation and explicit model selection provide complementary foundations for using 3D medical pretrained models. The benchmark supplies evidence about architecture, pretraining and task suitability that individual model reports cannot provide on their own. InsightRank connects this comparison to the cost of choosing candidates, demonstrating useful segmentation shortlists while identifying the need for better selection at the top of classification and regression rankings. Together, these contributions support evaluating medical foundation models by where they transfer well and how efficiently that suitability can be identified.

%% file: analysis/segmentation/segmentation_comparator_rows.tex
Same-indicator global vote & 0.4342 & 0.4798 & 0.6541 & 1/4 & 1.808 & 1.227 \\
Fixed cross-task recommendation & 0.3711 & 0.4901 & 0.6396 & 0/4 & 3.574 & 1.126 \\

%% file: sections/methods.tex
Our methods connect a common downstream benchmark with an evaluation of model selection before fine-tuning (Fig.~\ref{fig:benchmark_landscape}). We first define the candidate models, target tasks and fine-tuned reference rankings. We then describe the target-data samples and frozen representations used to calculate InsightRank and the baseline estimators. Segmentation uses voxel-level evidence, whereas classification and regression use one representation per patient. Finally, we specify the controlled comparisons, ranking and shortlist endpoints, and computational accounting used to assess the quality and cost of model selection.

\subsection{Benchmark design, model pool and downstream tasks}\label{sec:evaluation-design}
The benchmark compares pretrained models under a common set of downstream targets while retaining contrasts in architecture, pretraining objective and data composition. It contains 23 candidates---21 pretrained encoders and two architecture-matched scratch controls---evaluated on nine segmentation, two classification and three regression tasks. Each model--task pair contributes a three-fold mean, giving 322 reference outcomes. Tables~\ref{tab:model_pool} and~\ref{tab:tasks_summary} specify the model and dataset inventories.

The model pool uses OpenMind as a shared pretraining source for comparisons of architecture and learning objective. OpenMind contains approximately 114,000 three-dimensional head-and-neck volumes spanning several MRI sequences and derived image types~\cite{wald2025openmind}. Fourteen checkpoints use the full collection, two ResEncL MAE checkpoints use T1-weighted or image-quality-filtered subsets, and one DINO-style checkpoint uses OpenMind data. Five ResEncL--Primus-M pairs share the full-data source and registered objective; objectives can also be compared within each encoder family. The three ResEncL MAE variants provide a comparison of data composition under a common architecture and objective. BrainIAC, BrainSegFounder, SAM-Med3D and ViT-VoCo provide externally pretrained weights. ViT-VoCo inherits weights from the \href{https://github.com/Luffy03/Large-Scale-Medical}{VoCo repository}; CNN-VoCo and VoCo-SwinUNETR-L use OpenMind pretraining. External-weight comparisons therefore concern the released model configurations as a whole. The exact upstream checkpoint variant for ViT-VoCo was not independently verified.

Encoder groups are defined by network architecture for subsequent pairwise comparisons. Eleven candidates are ResEncL-like CNNs and twelve are transformers, comprising eight Primus-M, two ViT-UNETR and two Swin-UNETR configurations. The scratch controls use the corresponding downstream architectures with random initialization. Their group assignments follow their architectures, as for the pretrained candidates.

The task pool varies anatomy, imaging input and prediction granularity. Segmentation targets include lesions, tumours, cardiac structures, kidney/tumour regions and vascular structures; classification and regression use image-level diagnostic labels and continuous outcomes. Dataset counts distinguish participants, image series and analysed cases, and datasets are evaluated separately. The coronary-plaque and bridge-vessel CTA datasets are private hospital-curated cohorts. Plaque labels distinguish lumen, non-calcified plaque and calcified plaque, with annotations remaining under review. Bridge-vessel instance labels 1--5 are combined into one foreground class. That cohort contains 546 series from 507 patients; 44 series from 38 patients lacked matched masks and were encoded as all-background during training and evaluation. Its reported full-cohort results remain provisional.

\subsection{Downstream training and reference rankings}\label{sec:downstream-training}
Downstream fine-tuning supplies the reference against which model rankings and selection decisions are evaluated. Each model--task pair is summarized over folds 0--2. Segmentation is measured by Dice, classification by AUROC, and regression by mean absolute error. These task-specific outcomes are converted to a common ranking direction while retaining their original units for performance reporting.

Segmentation adapts the full pretrained network and its output layer to the target labels. Task-specific MRI or CT channels undergo nonzero-region cropping, channel-wise Z-score normalization and resampling to 1-mm isotropic spacing; ISLES and YBM use nonzero masks for normalization. Most candidates use $160^3$-voxel patches, with $96^3$ for BrainIAC and BrainSegFounder and $128^3$ for SAM-Med3D. Training runs for 150 epochs with 250 training iterations per epoch, using foreground oversampling and spatial and intensity augmentation. Dice plus cross-entropy is the usual segmentation loss. ResEncL runs predominantly use SGD with an initial learning rate of $10^{-3}$ and weight decay of $3\times10^{-5}$; Primus-M and most external-model runs use AdamW with $10^{-4}$ and 0.05, respectively. DINO3D uses an initial learning rate of $2\times10^{-4}$. The scratch controls use random initialization without loading pretrained weights.

Classification and regression use task-specific imaging modalities with binary-classification or single-output regression configurations. The recorded training settings specify $160^3$ patches, Adam with learning rate $10^{-4}$ and weight decay 0.01, and 100 epochs. Four-device training with one example per device and four gradient-accumulation steps gives an effective batch size of 16. Rotation and mirroring augment the inputs, and training uses mixed precision. Classification uses binary cross-entropy with logits; brain-age and survival regression use mean squared error.

Checkpoint selection and inference follow the protocol for each prediction type. Segmentation evaluates the final epoch-150 checkpoint using three-dimensional sliding windows with 50\% overlap, Gaussian weighting and mirroring test-time augmentation. Predictions are restored to the original image geometry, and each fold contributes its foreground-mean Dice. Classification and regression select the checkpoint with the lowest validation loss; MCI and IDH are evaluated by AUROC, and brain age, survival duration and post-stroke interval by mean absolute error. Original fold metrics are averaged with equal weight before ranking candidates.
% AUTHOR COMPLETION: Verify architecture-specific optimizer exceptions and learning-rate
% schedules, segmentation batch sizes/gradient accumulation, seeds and early stopping.
% Supply the classification/regression head construction and initialization, the exact
% post-stroke interval composite loss, and any output transformation or inverse scaling.

Reference rankings are calculated separately within each task. Let $\mathcal M_t$ denote the candidate set and $M_t=|\mathcal M_t|$. We define $Q_t(m)$ as mean Dice for segmentation, mean AUROC for classification and negative mean absolute error for regression, so that higher values consistently indicate better performance. Candidate ranks and architecture or pretraining contrasts use these within-task references. Raw outcomes with different units are not pooled across tasks.

\subsection{Target-data sampling and frozen-feature extraction}\label{sec:probes}
Transferability scores are calculated from fixed target-data samples shared by all candidates. For each task, a common manifest specifies the cases and labels used for feature extraction. Encoders run in evaluation mode with gradient calculation disabled. Segmentation retains features at labelled voxel locations, whereas classification and regression retain one feature vector per patient at each selected depth.
% AUTHOR COMPLETION: Data partitioning, patient-level separation and the associated
% verification protocol are deferred pending the author's revised specification.
% No new independence or overlap-exclusion claims are introduced in this draft.

Segmentation sampling preserves both class information and boundary locations within a bounded feature set. Deterministic class-stratified selection retains at most 100 patients per task and two cases per patient; the resulting samples contain 17--100 patients and 24--200 cases. Each case contributes at most 2,560 background points, 256 foreground points allocated proportionally to foreground class counts, and 200 additional ground-truth boundary anchors, with duplicate locations removed. The same coordinates are used for all models. Features from four approximately evenly spaced encoder depths are stored in FP16; the primary segmentation statistics use the final representation.

Voxel features are read from the encoder grid at locations corresponding to the sampled image coordinates. Images smaller than a patch are centrally padded, and larger volumes are processed in windows, assigning each sampled point to the first window that covers it. After accounting for padding and window offsets, coordinates are scaled by aligning the input and feature-grid endpoints and rounded to the nearest feature-grid location. Transformer patch tokens are reshaped into spatial grids where needed, and channel order is standardized before lookup. The extracted vectors retain their individual voxel identities rather than being averaged within a patient.

Classification and regression sampling preserves the distribution of the target while limiting the number of feature vectors. Selection retains one case per patient and at most 100 patients per task. Classification uses proportional class allocation; regression uses outcome quintiles for sampling allocation. SHA256 ordering with seed 20260915 and largest-remainder allocation determine the selected cases. Regression quintiles used for sampling do not replace the continuous outcomes in InsightRank's outcome-agreement statistic.

Image-level features summarize each patient's input by averaging spatial locations or patch tokens at four encoder depths. ResEncL, DINO3D and hierarchical Swin outputs use global spatial averaging; Primus-M and BrainIAC average patch tokens without using a CLS token. SAM-Med3D features are averaged over the spatial axes of the encoder-block outputs. Each patient contributes one forward input prepared for the encoder, with no additional averaging across multiple patches. Pooling applies no additional feature normalization; the subsequent statistics apply their own centring or normalization. The stored patient vectors are aligned by case identifier, retained in FP16 and converted to FP64 for calculation.

Depth selection follows the encoder structure. ResEncL uses stages 0/2/4/5, Primus-M uses blocks 3/7/11/15, and BrainIAC uses blocks 0/3/7/11. BrainSegFounder and VoCo-SwinUNETR-L use the four hierarchical layer outputs, while SAM-Med3D uses blocks 2/5/8/11 before its final neck. These selections sample positions along each network while preserving the native channel dimension of each representation.

Scratch feature extraction uses fixed random initializations without pretrained or fine-tuned weights. The segmentation factory defaults to initialization seed 0, and classification/regression extraction resets the initialization seed to 20261012 for each task and model. These scoring initializations were not verified to be identical to those used in the separately completed scratch fine-tuning runs.
% VERSION CHECK: Supplied AT-CGT-v3 I/P/Q and float64 graph/MST details describe a
% different segmentation method from the G/N/B InsightRank-v4.52 reported here.
% Do not merge these algorithm descriptions or change their associated results.

\subsection{Representation statistics for segmentation and image-level prediction}\label{sec:statistics}
InsightRank measures properties of frozen features at the scale of the prediction target. For segmentation, three statistics describe local class separability, activation amplitude and separation from a competing class. For classification and regression, two statistics describe the effective dimensionality of patient variation and agreement between neighbouring patients' outcomes. The definitions below specify each statistic's measurement and aggregation; their assignment to model pairs is given in the next section.

Class-balanced nearest-centroid accuracy, $G$, measures whether labelled voxel classes can be distinguished in normalized feature space. Let $x_i$ and $y_i$ be a sampled feature and its label. We set $z_i=x_i/\max(\lVert x_i\rVert_2,\epsilon)$ with $\epsilon=10^{-8}$ and normalize the mean feature of each class to obtain centroid $\mu_c$. For a case with observed classes $\mathcal C$, including background,
\begin{equation}
G_{\mathrm{case}}=\frac{1}{|\mathcal C|}\sum_{c\in\mathcal C}
\frac{1}{n_c}\sum_{i:y_i=c}\mathbf{1}\!\left[c=\arg\max_{c'}z_i^\top\mu_{c'}\right],
\end{equation}
where $n_c$ is the sampled count for class $c$ and at least two classes are required. Finite case scores are averaged within patients and then equally across patients to give $G_{t,m}$. Equal class weighting limits the influence of abundant background voxels. Centroids and queries are drawn from the same case, so this statistic describes local feature separability.

Activation amplitude, $N$, measures the magnitude of feature responses under the fixed extraction protocol. At most 16 feature vectors are sampled deterministically from each available patient--class unit $(p,c)\in\mathcal U_t$. With indices $I_{p,c}$,
\begin{equation}
N_{t,m}=\frac{1}{|\mathcal U_t|}\sum_{(p,c)\in\mathcal U_t}
\frac{1}{|I_{p,c}|}\sum_{i\in I_{p,c}}\log\max(\lVert x_i\rVert_2,\epsilon).
\end{equation}
Units receive equal weight, so patients with more observed classes contribute more units. Above the norm floor, a common positive rescaling $x\mapsto ax$ adds $\log a$ to $N$. The statistic consequently depends on feature scale, channel count and normalization placement; it measures response amplitude rather than class information.

Balanced centroid margin, $B$, measures the relative distance to the correct class and its closest competitor. Using $d(z,\mu_c)=1-z^\top\mu_c$, we define
\begin{equation}
b(z)=\frac{\min_{c\ne y}d(z,\mu_c)-d(z,\mu_y)}
{\max\!\left\{\min_{c\ne y}d(z,\mu_c)+d(z,\mu_y),\epsilon\right\}}.
\end{equation}
Eligible classes contain at least two sampled points, and each case requires at least two eligible classes. Margins are averaged within classes and equally across classes in a case. Finite case scores are averaged within patients and then equally across patients to obtain $B_{t,m}$. Larger raw margins indicate a stronger correct-class centroid advantage. Compared with the binary assignment in $G$, $B$ retains the relative angular distances to both centroids.

Effective dimensionality, $C$, measures how broadly variation between patients is distributed across feature directions. For model $m$ and depth $\ell$, we centre the columns of the patient-feature matrix $X_{m,\ell}$ and obtain eigenvalues $\lambda_j$ of its Gram matrix $X_cX_c^\top$. We calculate
\begin{equation}
\begin{aligned}
p_j&=\lambda_j/\sum_q\lambda_q,\\
E_{m,\ell}&=\exp\!\left(-\sum_jp_j\log p_j\right),\\
C_m&=\frac14\sum_{\ell=1}^{4}E_{m,\ell}.
\end{aligned}
\end{equation}
Negative numerical eigenvalues are clipped to zero, terms with $p_j\leq10^{-15}$ are omitted, and an eigenvalue sum below $10^{-30}$ gives $E=0$. Exponentiated Shannon entropy equals the number of nonzero directions when their spectral weights are equal and decreases as variation becomes concentrated in fewer directions. Thus $C$ measures the breadth of patient variation without using outcomes. The calculation uses covariance-proportional eigenvalues rather than unsquared singular values, with no PCA reduction.

Neighbour-outcome agreement, $R$, measures whether patients close in feature space have similar labels or continuous outcomes. Feature rows are normalized with a norm floor of $10^{-30}$, and distances are calculated as $d_{ij}=\max(0,2-2z_i^\top z_j)$. Setting $d_{ii}=\infty$ removes self-neighbours; stable distance sorting retains the 15 nearest other patients. For binary classification, agreement is the AUROC of the mean neighbour label. For regression, outcomes are converted to average-rank percentiles $u_i=(\operatorname{rank}(y_i)-0.5)/n$, and agreement is the Spearman correlation between $u_i$ and the mean neighbour percentile. Constant regression predictions receive zero. Averaging agreement over the two deepest representations gives $R_m$. Both $C$ and $R$ are invariant to a common positive scalar rescaling within numerical tolerances.

\subsection{Architecture-aware pairwise ranking}\label{sec:unified-rule}
Pairwise ranking combines the indicators through their orderings, allowing statistics with different units to contribute to one model ranking. Each encoder pair receives one indicator according to its architecture groups and the target type (Fig.~\ref{fig:benchmark_landscape}b). The fixed assignments are summarized below; a larger oriented indicator is preferred.
\begin{center}
\footnotesize\setlength{\tabcolsep}{3pt}
\begin{tabular}{lcc}
\toprule
Encoder pair & Segmentation & Classification/\\
 & & regression \\
\midrule
CNN--CNN & $G$ & $C$ \\
Transformer--transformer & $-N$ & $R$ \\
CNN--transformer & $-B$ & $C$ \\
\bottomrule
\end{tabular}
\end{center}

Indicator assignments and ranking directions are fixed empirical choices. For segmentation, the three pair types use class-balanced accuracy, negative activation amplitude and negative centroid margin, respectively. For classification and regression, transformer pairs use neighbour-outcome agreement and all other pairs use effective dimensionality. In particular, the negative direction of $B$ is a ranking choice established during development, whereas a larger raw $B$ denotes greater centroid separation. The primary rule is applied without fitting task-specific fusion weights.

Net-win aggregation converts pairwise comparisons into the final ranking. Let $a_i$ denote encoder $i$'s group, $q_t$ the target type and $h(q_t,a_i,a_j)$ the symmetric indicator assignment. Writing the oriented indicator as $s^{(r_{ij})}_{t,i}$,
\begin{equation}
\begin{aligned}
r_{ij}&=h(q_t,a_i,a_j),\\
v_{t,ij}&=\operatorname{sign}\!\left(s^{(r_{ij})}_{t,i}-s^{(r_{ij})}_{t,j}\right),\\
S_{t,i}&=\frac{1}{M_t-1}\sum_{j\ne i}v_{t,ij}.
\end{aligned}
\end{equation}
A win, loss or exact equality contributes $+1$, $-1$ or zero, respectively, with $v_{t,ji}=-v_{t,ij}$. Models are ranked by decreasing $S_{t,i}$, and the highest-ranked candidates form the shortlist. Strictly increasing transformations applied consistently to an indicator preserve its votes. The resulting score depends on the candidate pool: pairwise cycles can occur, and changing the number of models in each architecture group changes the balance of comparisons.

\subsection{Baseline transferability estimators and adaptations}\label{sec:baselines}
Baseline estimators are evaluated on the same target cases and candidate pool as InsightRank, using the declared representations for each task configuration. They summarize complementary aspects of the frozen features: LogME evaluates the evidence of a linear feature-to-target model, H-score compares between-class variation with overall covariance, and GBC measures overlap between Gaussian class distributions~\cite{you2021logme,bao2019information,pandy2022transferability}. NLEEP uses feature-derived mixture assignments to calculate expected empirical prediction, while CC-FV combines cross-case class consistency with feature variety for segmentation~\cite{li2021ranking,yang2023pick}. Their numerical implementations and adaptations are specified separately for the two prediction scales.

Segmentation comparators comprise CC-FV, LogME, GBC and H-score applied to the final encoder representation. Deterministic sampling caps each patient--class unit at 64 points and each task--class combination at 20,000 points. Randomized PCA with seed 0 retains at most 64 dimensions. CC-FV uses a scalable diagonal-Gaussian approximation to its distributional calculation, and GBC also uses diagonal class distributions. The reported CC-FV comparison therefore evaluates this approximation rather than the published full-covariance formulation.

The reported classification and regression comparison comprises LogME, H-score, spherical-Gaussian GBC and NLEEP. Unwhitened full-SVD PCA is fitted to the sampled target features and retains $\min(64,n-1,d)$ dimensions, where $n$ is patient count and $d$ is feature dimension. LogME accepts continuous regression targets directly; H-score, GBC and NLEEP use outcome quintiles calculated from the target sample as categorical regression adaptations. The principal comparison uses the deepest representation, with a four-layer mean additionally reported for LogME. NLEEP supplies Gaussian-mixture posteriors derived from features. Original source-posterior LEEP and NCE have no reported scores because source-class probabilities are unavailable. NLEEP is included in the image-level comparison only.
% BASELINE EXPANSION: TransRate-basic, PARC and 1-NN LOO have additional scores in
% classification_baseline_expansion_20261012/evaluation/. Their separate regression
% and PCA protocols and corresponding Results tables must be integrated together.

\subsection{Evaluation design and controlled comparisons}\label{sec:controls}
Controlled analyses distinguish the contribution of the indicator assignment from the quality of the component statistics and from simple model-selection strategies. All comparisons use the stated candidate pool and the same task-specific downstream references. The complete segmentation analysis summarizes nine tasks, and the predefined four-task comparator analysis comprises HNTS-MRG, KiTS19, CoronaryPlaque and CoronaryBridgeVessel. Classification and regression comparisons cover all five image-level tasks. The two coronary tasks retain the annotation qualifications described above.
% AUTHOR COMPLETION: Task-development and independent-evaluation boundaries,
% together with patient/data separation, will be specified in the deferred protocol.

The segmentation assignment control tests whether architecture-specific indicator selection improves on applying the same indicators to every pair. It keeps feature depth, indicator directions and net-win aggregation fixed, but assigns each pair a majority vote:
\begin{equation}
\begin{aligned}
v^{\mathrm{global}}_{ij}=\operatorname{sign}\!\bigl\{&
\operatorname{sign}(G_i-G_j)\\
&+\operatorname{sign}(N_j-N_i)\\
&+\operatorname{sign}(B_j-B_i)\bigr\}.
\end{aligned}
\end{equation}
Pair-type concordance additionally measures how often an indicator agrees with downstream ordering within each encoder-pair category. Predicted ties count as one-half agreement, downstream ties are excluded, and task-level concordances receive equal weight.

Classification and regression controls assess the contribution of the two image-level statistics and architecture membership. The controls rank candidates using $C$ alone, $R$ alone or a binary CNN-first prior. These comparisons help distinguish the information supplied by the feature statistics from ordering associated with the encoder group itself.

Budget controls compare the proposed shortlist with choices available without task-specific feature scoring. Uniform random selection samples candidate subsets of the same size, and its expected retained performance is calculated exactly. A fixed cross-task recommendation orders models by their mean rank across MSF, ACDC, YBM, ISLES 2022 and TopCoW, then applies that ordering to the four-task comparator analysis. Candidate-pool sensitivity analyses remove scratch models, restrict encoder groups or sample candidate subsets while keeping the primary rule fixed.

\subsection{Ranking accuracy and model-selection performance}\label{sec:evaluation}
Evaluation measures both agreement with the complete downstream ranking and the performance retained by a limited candidate shortlist. Ranking correlations assess the ordering of the model pool; Top~$k$ hit and regret assess the decision to fine-tune only $k$ candidates. All summaries give equal weight to tasks within the stated analysis, with segmentation and classification/regression reported separately.

Ranking accuracy is quantified by Kendall $\tau$-b, Spearman $\rho$ and weighted Kendall correlation. Segmentation uses SciPy's hyperbolic weighted Kendall coefficient as the primary rank endpoint, rounding both input arrays to ten decimal places before \texttt{weightedtau(score, target, rank=True)} with default additive weights. This implementation averages coefficients under decreasing lexicographic importance rankings in both argument orders and uses item weight $1/(r+1)$. Classification and regression use $\tau$-b and $\rho$ as principal endpoints and report the same weighted coefficient on unrounded inputs. Rank correlations retain ties; constant score vectors have undefined correlation.

Shortlist regret measures the performance lost relative to selecting the best candidate from the full pool. For predicted shortlist $A_{t,k}$,
\begin{equation}
\mathcal R_{t,k}=\max_{m\in\mathcal M_t}Q_t(m)-\max_{m\in A_{t,k}}Q_t(m).
\end{equation}
Top~$k$ hit records whether the shortlist contains a best reference model. Segmentation regret is reported in Dice percentage points. The classification/regression table reports Top~1 regret normalized by the observed task performance range, together with Top~3 hit; Top~3 regret is not available for that analysis. Thus the normalized image-level loss and the segmentation Dice-point loss are reported on separate scales.

Tie handling specifies how a ranking determines a finite shortlist. Segmentation uses stable descending predicted scores with ascending model IDs to resolve equal scores, independently of downstream Dice. Classification and regression average shortlist endpoints over uniform selection within a score tie at the cutoff; Top~3 hit is then the probability of including the best reference model. For uniform random subsets of size $k$, the expected best performance is
\begin{equation}
\mathbb E[Q_{\mathrm{best},k}]=
\frac{\sum_{j=k}^{M}Q_{(j)}\binom{j-1}{k-1}}{\binom{M}{k}},
\end{equation}
where $Q_{(j)}$ denotes increasingly ordered reference performance. Expected regret is the best full-pool performance minus this expectation.

Statistical summaries describe the observed benchmark and its declared comparisons. Three-fold means provide each model--task reference, and task means summarize each result panel. Candidate- and patient-subset percentiles, where analysed, describe repeat distributions. Individual model contrasts are not assigned confidence intervals because the folds are dependent and repeated fine-tuning seeds are unavailable. No confirmatory hypothesis tests are used for the reported comparisons.

\subsection{Computational cost analysis}\label{sec:cost}
Computational accounting compares exhaustive fine-tuning with extracting features from all candidates and fine-tuning only the selected models. We distinguish feature extraction, statistic calculation and downstream training so that the reported reduction refers to the operations actually timed. Job durations are summed over the included runs; they represent cumulative recorded job time.

Segmentation accounting uses the seven tasks with complete 150-epoch logs for all 23 candidates across folds 0--2. The latest completed retry supplies the duration for each model--task--fold run. Exhaustive fine-tuning sums 483 job records. Training the predicted Top~3 sums 63 records, to which all-candidate feature-extraction time is added. ISLES 2022 and HNTS-MRG are omitted from this timing analysis because seven required logs were unavailable. Feature-statistic computation was not separately timed and is excluded from the reported extraction-plus-training subtotal.

Classification and regression do not have a matched full-pool end-to-end timing comparison. Their ranking results therefore support no direct extrapolation of the segmentation time reduction. For the image-level algorithm, computational cost depends on the target-sample size and feature dimension: with $n$ patients and $d$ features, effective dimensionality costs $O(n^2d+n^3)$ per depth using the patient Gram matrix, and exact neighbour distances cost $O(n^2d)$ before row sorting. Pairwise aggregation costs $O(M_t^2)$ time and can accumulate net wins with $O(M_t)$ memory.